\documentclass[review]{elsarticle}
\journal{}
\usepackage[colorlinks=true, urlcolor=blue, pdfborder={0 0 0}]{hyperref}
\usepackage{natbib}
\biboptions{authoryear}
\usepackage[margin=2.5cm]{geometry}
\usepackage{array}
\usepackage{changepage}
\usepackage[ruled,vlined,linesnumbered]{algorithm2e}
\usepackage[numbered]{bookmark}
\usepackage{amsmath,amssymb,amsthm}
\usepackage{mathtools}
\allowdisplaybreaks[1]
\usepackage{tikz}
\usepackage{ulem}
\usepackage{longtable}
\usepackage{soul}
\usepackage{xfrac}
\usepackage{color, colortbl}
\usepackage{siunitx}
\usepackage{placeins}
\usepackage{textcomp}
\usepackage{bbding}
\usepackage{rotating}
\usepackage{multirow}
\usepackage{booktabs,siunitx}
\usepackage{array}
\usepackage{graphicx}
\usepackage{animate}
\usepackage{tikz}
\usepackage{lipsum}
\usepackage{float}
\usepackage{blindtext}
\usepackage{subcaption}
\usepackage{parskip}
\usepackage{soul}

\usepackage[margin=2.5cm]{geometry}
\usepackage{array}
\usepackage{changepage}
\usepackage[numbered]{bookmark}
\usepackage{mathtools}

\usepackage{ulem}
\usepackage{soul}
\usepackage{xfrac}
\usepackage{color, colortbl}
\usepackage{placeins}
\usepackage{textcomp}
\usepackage{bbding}
\usepackage{rotating}
\usepackage{multirow}
\usepackage{booktabs,siunitx}
\usepackage{animate}
\usepackage{algorithmicx}
\usepackage[noend]{algpseudocode}
\usepackage[thinc]{esdiff}
\usepackage[T1]{fontenc}

\DeclareSIUnit \year {year}
\DeclareSIUnit \kWh {kWh}
\DeclareSIUnit \tons {tons}
\DeclareSIUnit \rev {rev}
\renewcommand\appendix{\par
  \setcounter{section}{0}
  \setcounter{subsection}{0}
  \setcounter{figure}{0}
  \setcounter{table}{0}
  \renewcommand\thesection{Appendix \Alph{section}}
  \renewcommand\thefigure{\Alph{section}\arabic{figure}}
  \renewcommand\thetable{\Alph{section}\arabic{table}}
}
\theoremstyle{plain}
\theoremstyle{plain}
\theoremstyle{plain}

\usepackage[croatian]{babel}

\makeatletter
\def\UTFviii@undefined@err#1{??UPS??}
\makeatother
\begin{document}
\begin{frontmatter}
\title{The intersection of machine learning with forecasting and optimisation: theory and applications}
\author[Mahdiaaddress]{Mahdi Abolghasemi\href{https://orcid.org/0000-0003-3924-7695}{\includegraphics[scale=.6]{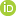}} }\ead{m.abolghasemi@uq.edu.au}

\begin{abstract}
Forecasting and optimisation are two major fields of operations research that are widely used in practice. These methods have contributed to each other growth in several ways. However, the nature of the relationship between these two fields and integrating them have not been explored or understood enough.  We advocate the integration of these two fields and explore several problems that require both forecasting and optimisation to deal with the uncertainties. We further investigate some of the methodologies that lie at the intersection of machine learning with prediction and optimisation to address real-world problems. Finally, we provide several research directions for those interested to work in this domain.
\end{abstract}


\end{frontmatter}

 
\section{Introduction} \label{sec:intro}

Forecasting (predictive analytics) and optimisation (prescriptive analytics) are two widely-used analytical techniques in academia and industry \footnote{In this study, we use forecasting and predictive analytics as changeable words. Similarly,  (constrained) optimisation and prescriptive analytic are used as changeable words}. While predictive analytics techniques are used to find the future values of a random variable, prescriptive analytics techniques are used for determining the optimal decisions given all the available information, i.e., to optimise objective function(s) with respect to some constraints and the existing uncertainty.

These two fields have been nurtured with decades of research and contributed to each other in several ways. For example, it is a common practice to estimate the parameters of regression models with optimisation models, i.e., by minimising the residuals. On the other hand, parameters of optimisation models are either predicted with some models or assumed to follow a certain distribution. The intersection of prediction, optimisation, and machine learning (ML) gives rise to several interesting problems that lack theoretical and empirical results. By leveraging ML models, we aim to explore two of them including i) \textit{predict and optimise}, ii) predicting the decisions of optimisation problems.

In the paradigm of \textit{predict and optimise}, we deal with problems that require sequential predictions and optimisations for decision-making. This paradigm is useful when the optimisation problem needs an input that is not fully observed but needs to be predicted. In this paradigm, we typically predict an unknown variable that will be used as an input parameter in an optimisation problem for determining optimal decisions. This paradigm is of interest from a forecasting perspective as it is concerned with a classic problem in forecasting literature whether forecasting accuracy translates to better decisions in downstream operations and how possibly predictions accuracy can impact the quality of final decisions. While often the objective of the problems that fall into this paradigm is not forecasting but the final decisions, from a forecasting perspective we are interested to know how forecasting accuracy from the first phase translates into decisions in the second phase. The literature suggests that better predictions do not necessarily lead to better decisions but there may be a correlation between them. The nature of the relationship between prediction accuracy and downstream decisions is still an open problem for researchers and practitioners \cite{abolghasemi2021effectively}. 

In the paradigm of \textit{predicting the decisions of optimisation problems}, the goal is to predict the values of decision variables for optimisation problems in real-time. Optimisation problems underpin rich mathematical backgrounds and have found numerous applications in solving real-world problems, e.g., scheduling trains and services, mixing ingredients, and determining facility locations, but their computational complexity hinders their real-time value and usage in practice. Any improvement in fast and reliable solving of these problems can be of huge benefit in practice.

One common approach to predict decisions is to collect training data offline (often solved by the mathematical model) and then present it to the model in supervised ML models. The availability of training data will not be an issue because one can solve them multiple times and collect data for training the ML models. However, there are some issues in this approach, the most important of which is ignoring the notion of constraints in the ML model that are inherent to the optimisation problem and encountering the danger of violating them. We will focus on the problems where predictive ML models can be used to estimate the decisions of optimisation problems while committing to the constraints or at least trying to abide by them. This paradigm is not investigated in the forecasting literature enough and lacks empirical and theoretical results, in general. 

In the following sections, we discuss the latest research concerned with the above-mentioned problems to open up an avenue for operations research researchers and practitioners who are interested to implement them or further exploring them.   There are tremendous opportunities that lie in the intersection of these two techniques and using them in conjunction with ML.


\section{Predict and optimise} \label{sec:predictandopt}

\subsection{Problem setting}\label{sec:problem1}
Forecasting is the basis for many managerial decisions that involve uncertainty. Forecasting per se may not be the end goal but is often a means to an end goal which is making decisions for a problem, whether that is determining the inventory level, scheduling some activities, or production planning. 

In the paradigm of \textit{predict and optimise}, the optimisation model takes inputs from the generated predictions by some model and accordingly determines an optimal solution given all the provided inputs. There is always a degree of error associated with these inputs (forecasts) and the optimisation models may accordingly change the prescribed solution with respect to the input values. However, it is not evident how forecast accuracy will be translated to different decisions in the optimisation phase where we aim to maximise or minimise an objective function.

This problem is important from both predictive analytics and prescriptive analytics perspectives. From a predictive analytics perspective, we are interested to know whether higher accuracy of predictions leads to a better decision \cite{abolghasemi2021state}. Generally, forecasters are interested to improve a model that not only predicts accurately on average but it is accurate over the horizon. While some forecasting and decision-making studies argue that forecasting is a separate matter from decisions and each of them should be modelled on their own, others argue that forecasts are not the final goal and should be used for better decision-making purposes, otherwise, there is no real benefit in them \cite{guo2013multivariate,goodwin2009common}. From prescriptive analytics, we are interested to develop models that can deal with the uncertainty in the predictions and find the optimal solutions for the entire decision-making horizon.

These two problems are typically modelled separately for a conventional \textit{predict and optimise} problem. If the predictions are near-optimal, i.e., close to the actual values, then separate prediction and optimisation should result in optimal decisions if the optimisation problem is modelled correctly. But this is seldom the case because the prediction problems can be complex. As such, the separate prediction and optimisation will be sub-optimal. Furthermore, technically the prediction model learns from a loss function that is typically not aligned with the optimisation model. So, the forecasting model ignores the final decisions in the learning process, i.e., its loss function is independent of the decision optimisation model. 

The integration of these two steps with ML can be beneficial to achieve a global optimal solution for the final decisions. Different techniques and methods have been proposed to deal with \textit{predict and optimise} problems. While there are no universal solutions for these problems, we explore some approaches that have shown promising results in practice.  

 

\subsection{Methodologies and applications}\label{method1}

The conventional approach for \textit{predict and optimise} is to solve them independently, i.e., generate forecasts with the model of choice and use them as inputs in the optimisation model to determine optimal decisions. Therefore, these approaches are more in the sequential form of `predict then optimise'. While this approach is feasible and may lead to good results it may not generate the overall optimal solutions. In this section, we discuss how these two phases can be integrated into a single integrated model to improve the results, i.e., forecast accuracy and decision quality. We advocate the \textit{predict and optimise} rather than \textit{predict then optimise} approach. We look at several different problems of this nature and explore how different techniques can be used to solve them. 

\citet{Bertsimas2020-li} proposed a model that goes from data directly to the decision. They develop models for a prescription where there is no constraint and show that we can improve the quality of decisions by leveraging the auxiliary variables that are available to the forecasting models but often are ignored by the decision-making model. In a seminal work by \citet{elmachtoub2022smart}, authors developed a customised loss function for smart predict and optimise (SPO)  where they included the objective and constraints of an optimisation model in the prediction phase. They introduced a decision-error function, i.e., the regret function,  that captures the decision errors measured by the difference between the decisions propagated by the predicted values and the actual values if they were known. Since calculating SPO is computationally expensive, they propose a convex surrogate loss function called SPO+, which is time efficient and works well with any polyhedral, convex, and mixed integer programming (MIP) problems with the linear objective. They discuss three different problems that are common in the realm of \textit{predict and optimise} in real-world: i) vehicle routing with a focus on the shortest path where the goal is to find the shortest path through a network of cities while the driver needs to go through all cities, each one only one time, starting from the origin and ending in the destination, ii) inventory optimisation where predicted demand is used as an input to find the optimal lot size or replenishment plan, iii) portfolio optimisation where the predicted price of stocks are used as inputs in an optimisation model to select a certain number of them, subject to the available budget or risk, and in order to maximise the profit. However, the authors run computational experiments only on the shortest path problem and portfolio optimisation problems and show that the SPO+ is consistent with the least squared and is superior to other conventional loss functions when the model is misspecified. 

\citet{mandi2020smart} looked at the smart predict and optimise for hard combinatorial optimisation problems. They further explore the SPO and SPO+ in discrete optimisation problems where the objective is not linear. Such a problem is computationally expensive since one needs to solve the optimisation problem sequentially. They relax the optimisation problem, i.e., linear problem instead of a MIP problem, and explore how warm start learning and relaxed optimisation can improve their results. Warm start learning in this context refers to the learning of the predictive model first from the standard mean squared error loss function and then learning from the regret function. Their empirical results on the Knapsack problem and Irish electricity data show that the relaxed optimisation problem achieves similar results to the full discrete optimisation problem. In terms of computational time, warm-start learning speeds up the learning at early stages but does not depict a long-term gain. While SPO shows some promising results on small discrete optimisation problems, they are not tractable on large problems. 

In an interesting work by \citet{elmachtoub2020decision}, authors trained decisions by SPO trees loss function which was inspired by SPO and showed that the new decision trees have a higher accuracy and decision quality in comparison to conventional tree-based models. Their model is more interpretable and provides a base for using tree-based models to predict and optimise problems. 

\citet{wilder2019melding} raised the main drawback of the \textit{predict and optimise} problems, i.e, treating these two phases independently (two-stage models) and training them via different loss functions that may not be aligned. They then proposed an integrated end-to-end model for decision-focused learning where they trained a deep learning model that directly learns the decisions.  They focused on two classes of combinatorial optimisation problems including linear programs and submodular maximisation problems, both of which have many real-world applications, e.g., finding the shortest path with linear programs, and recommendation systems with submodular maximisation. The challenge for integrating the two phases when the optimisation models are not convex lies in the nature of solving these problems. While prediction models are learned with gradient descent, the optimisation models may be non-differentiable. To remedy this problem, they propose to relax the optimisation problem to a differentiable continuous problem and derive a  function that can be used as a continuous proxy for training the integrated model. They empirically test their proposed methodologies on matching, budget allocation, and recommendation problems all of which have an element of prediction and an element of optimisation. They use neural networks for decision-focused learning and compare them with two-stage approaches. The results show that while the accuracy of two-stage models is often better than the decision-focused ones, the quality of solutions for decision-focused models is significantly better than the two-stage models. One interesting finding of their research is that the correlation between the predicted values of the two-stage neural network with ground truth is lower than the ones for decision-focused models, even though the two-stage models outperform the decision-focused models in terms of accuracy. This shows that the decision-focused models can learn better and account for outliers to obtain a better decision overall.

In another study, authors integrate the prediction and optimisation models by learning losses through specific tasks rather than using the surrogate loss function \cite{shah2022learning}.   They propose locally optimised decision loss (LODL) and argue that their model learns the decisions automatically via their proposed methodology as opposed to other integrated models that use a hand-crafted surrogate function for learning the decisions. Suppose for a given set of features $x$,  a predictive model $P_\theta$, generates forecasts $\hat{y}$. The predictions will be used to find the optimal decisions $z^*$ in the optimisation model $\text{argmin} (f(z,\hat{y})) ~ \text{s.t.}~ g_i(z)$ $\leq$ 0, for $i \in {1, 2, ..., m}$.

In decision-focused models, decision loss $dl(\hat{y},y)$ can be computed via $(z^*(\hat{y}, y)$. That is, for a set of n features and labels $(\textbf{x},y)$, we train a predictive model $M_\theta$ that minimises the dl, $argmin_\theta$ $\sum_{i=1}^{n}$ dl($M_\theta(x_i),y_i)$. In this setting, the predictive model is using decision loss for learning as opposed to conventional forecasting metrics. In LODL, the authors estimate a new parametric loss function $LODL_\phi(\hat{y}, y)$ that approximates the dl behaviour and is convex by nature. So, one can train the model with the new loss function. They empirically show that their model works well on three domains of problems including portfolio optimisation, web advertising, and linear models.

\citet{mulamba2020contrastive} introduced a new class of loss functions that were inspired by noise contrastive estimation and used inner approximation along with a look-up strategy to speed up the computation. Essentially, they proposed a new loss function that is differentiable and does not require solving the optimisation model. Based on this work, \cite{mandi2022decision} proposed to look at decision-focused models as a learning-to-rank problem that aims to learn the objective function while ranking the solutions in terms of the overall objective. In this setting, different surrogate loss functions can be used for ranking the solutions. The authors used point-wise ranking which looks at each feasible solution independently, pair-wise which looks at the best versus rest, and list-wise ranking which looks at the orders of solutions. The empirical results on the shortest path, energy scheduling, and Bipartite Matching shows some promising results although not always superior to other existing methods in the literature.  

In a study by \citet{demirovic2019investigation}, they looked at the \textit{predict and optimise} problem for the knapsack problem. The knapsack problem is a classic combinatorial optimisation problem that includes prediction and optimisation problems. In Knapsack, the decision maker should select a number of items from a set of $n$ items each of which has a known weight of $w$ and profit of $p$ which may be known or not, such that the total value of items is maximised and the total weight of the items does not exceed a fixed size $c$. Knapsack can be formalised as follows:
\begin{align}
max \sum_{i=1}^{n} p_i x_i\\\nonumber
\text{subject~to}:
\sum_{i=1}^{n} w_ix_i \leq c, \\\nonumber
x_i \in {0,1} 
\end{align}\label{knapsack}

In this problem, if the profit of an item is not known, then we need to predict it. Given that the prediction will have some associated uncertainty and error, the optimisation model may prescribe a different solution depending on the predictions values and errors. In this problem, the end goal is to minimise the error in optimisation model. One way to do so, is to minimise the regret of predictions in the optimisation, which is the difference between the optimal solutions with actual values of $p$, and prescribed solution with predicted values $\hat{p}$. This is challenging if the regret function is not differentiable and one may use a surrogate loss function to minimise that. They compared indirect, direct, and semi-direct approaches. Indirect approach is simply predicting and optimising individually and independently, direct approach uses a convex surrogate of regret as loss function and solves the optimisation problem at each time stamp, and semi-direct approach uses a convex surrogate regret function that uses the parameters of the optimisation model but does not need to solve that repeatedly. For each of these approaches, they use different methods for learning and test their methods on two different dataset, one synthetic data and one energy pricing data. They report while there is limited benefit in including optimisation model for the synthetic data, there may be more benefit for the energy data which is more noisy. For the investigated methods, they state that there is a potential in direct and semi-direct methods and they may outperform the indirect methods when the optimisation model is near convex. The indirect methods may be more appropriate for difficult optimisation problems, and more studies are required to reach a more concrete conclusion.

 In a different setting to above-mentioned examples, prediction and optimisation have been explored in stochastic programming problems.  Stochastic programming is commonly used for decision-making under uncertainty.  If the distribution of inputs is known but the objective cannot be defined and solved analytically, one can use Monte Carlo to draw samples and treat the model as a deterministic model. If the distribution is not known, one can empirically estimate their distribution from samples. This is an active research topic. The traditional approach is to use sample average approximation (SAA) where the actual distribution of a variable is replaced by an estimation.  There are several other methods including robust optimisation, robust SAA, and stochastic optimisation \cite{bertsimas2019machine}. ML has been used in various forms for solving stochastic programming problems. For example, \citet{donti2017task} proposed end-to-end learning in stochastic optimisation problems. They investigated probabilistic ML models that produce predictions by considering the objective of the stochastic optimisation models.  They intended to use the predictions in the stochastic programming model in a way that they lead to best results for the stochastic programming model but not the most accurate forecasts. They looked at three problems in energy price forecasting and battery storage, load forecasting and generator scheduling, and a synthetic inventory control problem. The empirical results show that the task-based learning where they have learned the predictions for tasks outperforms the conventional stochastic programming methods. \citet{larsen2022predicting} also looked at the two-stage stochastic programming problem in a similar setting, but their goal was to rapidly predict the tactical solutions in the second stage as they are computationally demanding. They formulated the problem as a stochastic programming problem and predicted the solutions via ML models. Their empirical results on load planning for transportation shows that the predictive accuracy  of deep learning models is close to the lower bounds of the stochastic prediction programs (to be read and completed)? that is calculated based on sample average approximation.

These are some of the main works in the domain of \textit{predict and optimise}. However, there are other studies that have looked at the same problem with different techniques like constrained predictions. Fore more details, we refer interested readers to a review paper by \citet{Kotary_undated-mq}.


\subsection{Discussion and future works}\label{discussion1}

There is a long-standing debate in forecasting literature that looks at the value of predictions or the utility of predictions. That is, what is the final goal of forecasts and how they may be used for decision-making \cite{abolghasemi2021state}. This problem has been investigated from both psychological and judgemental aspects \cite{goodwin2009common}, as well as mathematical and statistical aspects \cite{abolghasemi2021effectively}. Forecasting can be used as a separate task that is independent of decision-making. As such, decision-makers can have their interpretation of forecasts and take action accordingly. One may argue that if the purpose of predictive models is to use them for decision-making, we should incorporate the elements of decision-making in the predictive models directly \cite{bertsimas2019machine,Wilder2019-wq}. But it is not always straightforward to integrate some elements of decisions in the forecasting. This may be due to the complexity of the decision-making model, e.g, if we have a complex objective function that is not differentiable, or the objectives in decision-making phase have subjective nature and can not be modeled mathematically. For example, it is not evident how one can embed non-monetary objectives or judgmental elements in the decision-making process and optimise them. 

The spectrum of the \textit{predict and optimise} problems is vast and there is no unique solution for them. Depending on the type of prediction and optimisation problem, the number of constraints in the optimisation model, and the final goal, we may be dealing with convex and non-convex problems which require different approaches. Most of the \textit{predict and optimise} methods have been proposed and tested on optimisation problems with a linear objective. The empirical results on linear programming are promising but generally lacking on broader aspects of the optimisation problems \cite{Donti_undated-sq}. The optimisation model may have a discrete and non-differentiable function or complex objective function with complex constraints. This will make the problem difficult and in some scenarios computationally expensive. Extending these methods and verifying them on MIP problems is an interesting research avenue with numerous real-world applications. 

 The computational time is an important barrier in implementing some of the problems. It would be useful to find scalable solutions to these problems. This can help to use them for solving problems in real-time.  In next section, we discuss the idea of predicting the solution of optimisation models with surrogate ML models which can help to speed up the process but there is a need for more rigorous studies in this domain. 

As discussed, one way to integrate decisions in forecasting is to use a  surrogate loss function in the model. These surrogates are often estimated by adding regularisation terms to the objective function or relaxing the optimisation problem. It is not trivial how one can build these surrogate functions. The main drawback of this approach is that one needs to design a surrogate loss function for each problem individually. There is no rigorous method or algorithm to design them and it is more of an art that is dependent on the problem at hand. Developing a general-purpose algorithm and guidelines, at least for some types of problems, would be useful. Furthermore, no software or package integrates these two phases of prediction and optimisation. While it may not be easy to develop general-purpose software that covers all types of prediction and optimisation models, we strongly believe that integrated software for simple cases of prediction and linear optimisation could be beneficial.

Another important problem that is worth investigating is concerned with modelling the uncertainty of forecasts and optimisation. We want to make forecasts more accurate where it matters the most, and not just on average. This can be done by incorporating the information from the decision optimisation phase. However, there is no rigorous study to look at the distribution of forecast errors and examine how they translate to optimisation costs in downstream operations. While there is empirical evidence that the prediction accuracy will impact the optimisation cost \cite{gitmahdi}, it is not evident whether and how the optimisation model will be able to deal with forecast misspecifications. Empirical and theoretical analytics will be of tremendous value in this domain.

There is no standard benchmarking method or dataset to develop and evaluate \textit{predict and optimise} methodologies. Knapsack problem \cite{mandi2020smart} could be a good candidate but that is limited to its form of objective and constraints and thus the metholdogies may not be applicable for other \textit{predict and optimise} problems. Several studies have looked at this problem for different problems such as optimal power flow \cite{chatzos2020high}, and scheduling \cite{abolghasemi2021state}, but the applications are not limited to these problems and one can use these methods for numerous other areas.  Some interesting application is portfolio selection where stocks price should be predicted and a set of them selected for maximum profit similar to M6 forecasting competition, production planning where demand for a product needs to be predicted and then accordingly production needs to be scheduled, staff scheduling where job demands should be predicted and then used in an optimisation model for scheduling staff, just to name a few. Although there are some studies in these domains, the empirical results are lacking.

\section{Predicting the solutions of optimisation problems}\label{sec:predictopt}

\subsection{Problem setting}\label{problem2}

We first provide a general introduction to constrained optimisation models before looking at various methodologies and applications of ML in constrained optimisation. Constrained Optimisation problems are concerned with finding the optimal solutions for a problem with a set of constraints. Constraint optimisation problems aim to minimise a function $f$ of variables $x \in R^n$, subject to a set of constraints $C$ that must be satisfied:
\begin{align}
&\min_{x}f(x) ~\\ \nonumber
& \text{subject~to:} x \in C. 
\end{align}
If a solution $\eta$ satisfies the constraints, it is called a feasible solution, and if $f(\eta) \leq f(Z) $, for all feasible $Z$, then it is an optimal solution. These problems have found numerous applications in practice and are celebrated for their strength in finding the best or optimal solutions. Some examples include how to schedule staff or cars for a given number of tasks by minimising the total costs, or how to combine some ingredients to satisfy the requirements of a product while maximising the profit. 

The optimisation literature benefits from a rich mathematical background and the community have advanced these methods during the last decade significantly. The constrained optimisation problems may be solvable efficiently or sometimes it may be hard to find a tractable solution for them.  If the problem is convex, i.e., the objective function is a convex function and the solutions space is a convex set, then there are efficient methods that can find the optimal solutions for the problem. One common case is linear programming problems, where the objective function and constraints are linear for which we can find their optimal solution efficiently. However, if one of the variables needs to be an integer, we may not be able to find an efficient or optimal solution. These problems are also called, combinatorial optimisation problems, a particular type of constrained optimisation problems that are characterised by their discrete space and combinatorial nature of solutions. If any of the constraints or the objective function is non-linear, we call them nonlinear programming again these problems are hard in nature and they may not be tractable. 

In general, the intersection of ML and optimisation can be explored from two different perspectives, one being the contribution of optimisation to ML, e.g., finding the optimal decision tree with constrained optimisation models \cite{bertsimas2019machine, Daume_undated-vo, Khalil_undated-qm}, and the other being the contribution of ML to optimisation models, e.g., finding the solution of optimisation with ML models \cite{abolghasemi2021effectively,bertsimas2019machine}.  While this can be further classified, without loss of generality, the focus of this study is on the latter.

The main motivation of using ML models for predicting the solutions of optimisations problems is their computational speed. Optimisation problems are often computationally expensive and it may not be practical to use them for real-time decision-making. ML models are the most popular methods for predicting the decisions of optimisation problems. The possibility of customising their loss function, and the non-linear nature of many ML models like neural network makes them suitable choice for predicting the solutions of complex optimisation problems. This will enable us to use them for real-time decision-making.

\subsection{Methodologies and applications}\label{method2}
 
The intersection of ML and optimisation problems are heavily influenced by understanding how ML models can help optimisation models find their solution faster and in a more systematic way. One such contribution of ML to optimisation is to facilitate finding the solutions of optimisation problems, in particular, to find the branching variables in solving MIP problems via using `Branch and Bound' algorithm \cite{bertsimas2017optimal}. Finding the branching variables can significantly improve the speed of solving MIP problems.

'Branch and Bound' algorithm, seeks the best feasible solutions for MIP optimisation problems by enumerating the possible feasible solutions. After constructing the root node which consists of the entire search space, the algorithm decides which node it is going to explore next. This is based on the internal check where the optimal solutions are calculated and if there is no benefit in exploring the node, the algorithm will discard it.  The branching is done in rather a parsimonious way and although there has been a lot of growth in better choosing these variables and also speeding them up, there is no theoretically guaranteed solutions for them.

Typically branching methods can be classified into two main approaches: i) Strong branching (SB) which tests variables at each node and choose the best one that can improve the solution from its current state and decrease the gap to optimally, ii) Pseudocost branching (PB) that mimics the behaviour of SB but speeds up the computation by looking at a smaller number of variables for branching. 

Many different methods have been proposed for branching but most of them rely on manual feature engineering, heuristic and ad-hoc methods. ML provides an alternative for a more systematic way of branching.  There has been a great interest and a number of studies that explore the application of ML in branching. Various techniques such as supervised classification \cite{marcos2015machine}, regression \cite{Daume_undated-vo}, and ranking \cite{Khalil_undated-qm} methods have been successfully used for this purpose. For example, \citet{bertsimas2017optimal} developed a discrete optimisation model to obtain the optimal branching in decision trees. Decision trees essentially branch on features in a recursive manner to find the best possible solutions. The branching is done using a greedy search where the model selects a feature to find a local optimal. There is no guarantee that the constructed tree will lead to optimal results. They developed a mixed-integer model and tested it on 53 datasets with univariate and multivariate settings. The empirical results showed an improvement of 1-2\% for univariate models and 3-5\% for multivariate models in comparison to the classification and regression trees algorithm. This accuracy depends on the depth of the tree, and the gain in out-of-sample accuracy is often larger for smaller depths.

Another application of ML in optimisation, as discussed previously, is to find the solution of optimisation models directly rather than facilitating the process to find the solutions. This is useful because ML models are often significantly faster than the optimisation models and if we can train an ML model to predict the solution of optimisation problems, we can use them for real-time decision-making. Note that the notion of time is a relative concept when we refer to solving optimisation problems. Traditionally MIP problems are known to be intractable. Although the algorithms developed by researchers, CPLEX, and GUROBI are significantly faster in comparison to a decade ago, in some cases a million times faster \cite{bertsimas2019machine}. Not all MIP problems are computationally expensive but it is not uncommon to find NP-hard problems (there may not be an algorithm that can solve the problem in time that is polynomial in the size of inputs) in this class of problems.


How does the process of using ML for predicting the solution of optimisation models work? One can run the optimisation model several times to create a large training dataset. Then, use the set of inputs including parameters, and outputs including all decision variables to train a supervised ML model.  Such a model will train based on mapping inputs to outputs according to the loss function of choice. Multi-class classification \cite{bertsimas2019machine} and multi-output regression \cite{abolghasemi2021effectively} have been proposed for this purpose. The choice of the loss function is important for the successful implementation of these algorithms. Ideally, the loss function should be customised according to the objective function of the optimisation model but this may not be always possible.

The benefit of using supervised ML models is their significantly fast performance in comparison to the optimisation models, making them the best choice when real-time decision-makings required. While optimisation models are not always slow, they are still lagging behind most ML algorithms in terms of computational speed.  

Several problems should be treated with caution when using an ML model for predicting the decisions of constrained optimisation models including i) committing to constraints, ii) requiring sample data for learning, iii) the ability to generalise.

First, ML models generally are unconstrained optimisation models and one can not guarantee they will commit to the constraints, whether it be physical or business constraints. Some studies attempt to develop a constrained predictive model to estimate the solutions of optimisation models \cite{chatzos2020high}. Several methods have been proposed to consider the constraints in ML models while learning. Perhaps the most promising approach is to use Lagrangian duality \cite{Fioretto:ECML20}. The Lagrangian method is a promising method in optimisation to solve complex non-convex problems. For many primal non-convex problems we can build a  dual form with Lagrangian. Suppose an optimisation problem

\begin{align}
& \min_{}f_0(x)\\\nonumber
& s.t: f_i(x) \leq  0, ~~ i = 1, \ldots, n.\\\nonumber
& g_j(x)=0, ~~ j = 1, \ldots, m.
\end{align}\label{primal}
Then, the Lagrangian relaxation function for this optimisation problem is 
\begin{align}
\min ~ L(x,\lambda) = f_0(x) + \sum_{i=1}^{n} \lambda_i f_i(x) + \sum_{j=1}^{m}  \lambda_j g_j(x) , \\\nonumber
\end{align} 

where $\lambda_i$ and $\lambda_j$ are called Lagrangian multipliers. The violation-based Lagrangian can be expressed as 

\begin{align}
\min ~ L(x,\lambda) = f_0(x) + \sum_{i=1}^{n}  \lambda_i {f_i(x)} + \sum_{j=1}^{m}   \lambda_j max(0,g_j(x)) , 
\end{align} 
In fact, $\lambda_i$ and $\lambda_j$ measure the violation degrees for the constraints.

\citet{Fioretto:ECML20} used Lagrangian violation-based relaxation in AC optimal overflow problem in power systems. They developed a deep neural network architecture where they adopted the loss function to account for the physical and engineering constraints. They tested their model on empirical real-world data and showed that their model largely satisfies the constraints and generates reliable solutions significantly faster. \citet{fioretto2020lagrangian} proposed a  Lagrangian dual approach to deal with constraints in deep learning models.  They applied their model to energy data and showed that the proposed model is two orders of magnitude better than the traditional deep learning models that use mean squared error as the loss function. \citet{tran2021differentially} used Lagrangian duality and adopted neural networks to address the problem of fairness, accuracy and privacy in supervised learning problems where data may be imbalanced and learning algorithms may not have all the information due to privacy concerns for proper learning, thus discriminating against some groups.  \citet{chatzos2020high} showed how ML models can be used to approximate the solutions of large-scale optimisation problems in power flow settings. They proposed deep learning models with the Lagrangian duality that can achieve the solutions within 0.01\% of the optimality in milliseconds while capturing the constraints with high fidelity. Similarly, \citet{donti2021dc3} proposed a deep learning architecture called Deep Constraint Completion and Correction (DC3) which enforces feasibility and is differentiable. The empirical results on synthetic data and power flow show near-optimal results while committing to the constraints. While the loss in optimally is marginal, the computational speed makes them a viable choice when dealing with problems that require instant decisions.

The above-mentioned models generally consider the constraints in a soft-manner, i.e., we should introduce the constraints and they may still violate the constraints depending on how we set them up. However, they do not learn the constraints. This may limit their benefits as they are unable to intelligently learn the constraints. There are some solutions in the literature that suggested overcoming this limitation. Deep neural networks and graph neural networks are two of the widely used techniques to incorporate the constraints of the optimisation models. For example, \citet{detassis2021teaching} proposed a deep learning model where the models were able to learn the constraints by supervised ML models. This is different to previously discussed models where constraints were not learned but only considered in implementation. They tested their model on a relatively simple constrained model and showed promising empirical results. However, their results are approximate and not exact in terms of committing to the constraints and obtaining final solutions.

Constraint programming is another methodology that can be used to account for constraints and solve optimisation problems. Although constraint programming can be used for solving combinatorial optimisation problems, we differentiate this application. Constraint programming fundamentally relies on logic programming and graph theory from computer science, but it has also benefited from mathematical programming in developing methodologies. A common application of constraint programming is to use it to find heuristic solutions for MIP problems. Constraint programming works by reducing the solutions space until all the decision variables meet all the constraints and then values are assigned to decision variables. ML techniques have been used in optimising the search process and also predicting their solutions. While constraint programming is a paradigm that is used for solving MIP problems, it is a different problem on its own and its solutions can be predicted with ML. Similar to MIP or other optimisation problems, constraints can be considered in different ways. For example, constraint violations can be considered as penalties in the objective function. This may not guarantee that the solution commits to the constraints but it can facilitate that. Another technique is to hard code the constraints in the ML models by enforcing them to fall between the boundaries. See \cite{Popescu2022-tk}, for a comprehensive review of the applications of ML in constraint solving.

Second, ML models need a set of training data to learn the process. This means we need to first develop an optimisation model and run it to obtain the required outputs. This way the performance of ML models always depends on the performance of the optimisation model and if the quality of the solutions for the optimisation model is poor, ML model will perform poorly accordingly. Having said that, if the optimisation model is not solved to the optimality or heuristics are used to approximate their solution, then ML models may be able to learn the process better and outperform them marginally. However, if the optimisation models are solved optimally then ML models almost always will underperform them.

Third, surrogate models may not generalise. If there is any change in parameters or setting of the problem in objective or constraints, the surrogate model may not work anymore. One needs to run them through another model to obtain the inputs and outputs, and retrain a surrogate ML model to learn the mapping process again. There is no empirical or theoretical evidence to show how much the performance of surrogate models will change. The problems also exist when we have different input and output lengths, although some solutions have been proposed to deal with this problem \cite{vinyals2015pointer}.
If the model is used for one specific problem, then this is no longer an issue because the model can be developed and solved only once. For example, a network of a blood supply chain in hospitals may not change the business setting, and once the model is developed and trained it does not to be changed anymore or at least for a while. However, if there is a different dynamic then models may be needed to change more frequently and the surrogate model may not be a good choice.  RL is an alternative that has been used to overcome this issue.

One can combine reinforcement learning (RL) with the supervised learning problem, to improve the performance of the surrogate model, enabling it for better generalisation, and commit to the constraints. Almost all of the RL problems can be formulated via Markov Decision Processes (MDP). We can define MDP with the vector $M = <S,A,R, T, \eta,H>$ where $S$ is the state space comprising the initial and possible solutions for a problem, $A$ is the set of actions that an agent can take, $R$ is the reward function that dictates how the actions will improve or decrease the values for a particular solution, $T$ is the transition function that governs the transitions from one state to another in a conditional form $p(s_{t+1} s_t,a_t)$, $\eta$ is a discount factor which takes values between zero and one whose goal is to encourage for short-term rewards, and H is the length of actions performed to reach the solution. In mathematical terms, the goal is to maximise the reward, $\max_{}E[\sum_{t=0}^{H} \eta_t R(s_t, a_t)]$. Agents in RL algorithms look for an optimal policy that maximises the reward. This can be done either by i) maximising the value action function which is the expected reward for a policy given a state and an action, or ii) directly modelling the agent's policy as a function and optimising it to maximise the reward.


The policy can be learned from the demonstration as done in imitating learning. The idea of providing a sample behaviour for training the model is also known as imitation learning. In imitation learning, a machine is trained to learn a task after being presented to an expert behaviour (e.g., human or sensors), therefore, learning to map inputs to outputs. Machines can also learn from experiences, where the experience can be defined by an MDP, which shows the agent's state, action, reward and new state after taking an action. Imitation learning is useful for learning complex tasks as it provides the opportunity to demonstrate the behaviour to a machine without having the exact policy function in place, simplifying the learning process. It has many applications including robotics, self-driving cars, and games to name a few. In these tasks, the machine is supposed to make a decision and take an action given its current state. This may be doable by an optimisation model. However, the number of scenarios over a horizon often becomes too large, making an optimisation model obsolete. Moreover, it is not trivial what the loss and reward function is and how one can define them, making optimisation models impractical for these problems. As such, often human behaviour is shown to a machine (agent) for enabling them to directly learn from experts. The challenge is the generalization to unseen states in future because we cannot demonstrate all the behaviours to an agent, but show only one specific behaviour that may not be even optimal. It is useful t combine them with RL and optimisation techniques to optimise the policy. For a review of the current state of the art and future challenges, see \cite{Hussein2017-tb}

In applying RL to an optimisation problem, we define an environment and agents whose aim is to find optimal solutions. The agents and environment interact in a setting where the goal is to maximise the total cumulative agents' rewards with respect to a policy. The policy is often stochastic and the agent may get a reward or penalty depending on the action it takes. RL starts with a random policy and improves this policy over time by learning through trial and error actions.  The ecosystem also includes a value function which is the expected return for a particular action, given the current state of the agent. The goal is to build an optimal policy, i.e., an optimal loss function. For this, the agent on one hand may try exploring new states for possible rewards and on the other hand, repeating the actions that it knows or predicts will return a reward, all with the aim of maximising the total reward. This is known as exploration vs exploitation. All these actions take place in an environment that follows the Markov process,i.e., the future state of the system only depends on the current state and it depends on the older states only through the current state. The agent learns from the experiences and decides on a given MDP. After taking the action and moving into a new state, the agents will update the parameters and keep repeating the process until the end of the horizon. RL can be used to optimise the parameters of the policy. Imitation learning via demonstrating the experts' behaviour was shown to agents to speed up policy learning. The most well-known application of combining imitating learning and RL is in the game `Go` where a recurrent neural network was used for imitation learning from previous experiences and RL was used during the game for refining the weights and adopting the behaviour \cite{Silver2016-wn}. 

RL has been used to solve various optimisation problems. Perhaps the most famous application is Travelling Sales Man (TSP). TSP is a combinatorial optimisation problem in which for a set of given cities, a traveller must visit each city exactly once while minimizing the total distance. For TSP, we can define an MDP as follows: the state is the cities that the traveller has visited so far with the initial state being the starting point, actions are the city that the traveller can visit, the reward is the negative of the tour length, and the transition function is the chain of nodes until all of the nodes have been visited. \citet{bello2016neural} and \citet{Dai_undated-wp} are some of the early researchers who proposed to use a graph neural network to sequentially select the nodes(cities) on the graph. The selection policy was learned with RL. This approach proved promising and other studies have adopted similar approaches. See \cite{cappart2021combinatorial,Mazyavkina2021-tl} for a review of graph neural networks in optimisation and RL application in other typical optimisation problems.

\subsection{Discussion and future works}\label{sec:discussion2}

The methods and applications of ML in predicting the decisions of optimisation problems lack theoretical and empirical results. As discussed various ML models such as deep learning, graph neural network, and tree-based model have been used for predicting the decisions of optimisation problems \cite{abolghasemi2021effectively}. There is no consensus on which algorithms may be more useful for which type of optimisation problems. The problems are also largely unexplored in the configuration of the ML models, e.g., what type of neural network architecture with what setting of hyperparameters would be more useful.  

We previously discussed the possibility of mapping optimisation inputs to outputs (decision variables) via ML models, just like a typical supervised learning problem. The drawback of these techniques is that they typically ignore the constraint, if there is any. One interesting research avenue is to use constraint learning techniques to learn the constraints from samples. The current techniques can be used to learn soft constraints where there is a preference for constraints, or for learning hard constraints such as Boolean constraints \cite{de2018learning}. This can be naturally extended to learn from solved samples for a typical optimisation problem. The feasibility of constraints is a big challenge in predicting the solutions of optimisation models. One can use deep learning models with a final layer to map the solutions into a feasible region. 

When using ML to predict the solutions of optimisation, one critical point that we should consider is the scale of outputs and the metrics that are used in surrogate modelling. Surrogate models use another model, often simple, with another evaluation metric, to estimate the solutions of an original problem with a different metric. The evaluation metrics may be of a completely different nature, and since the goal of surrogate models is to minimise the error of the original model, it is not enough to evaluate their performance based on their metrics, rather the original metrics should be taken into account. A promising approach is to estimate the actual loss function and use that in the ML model to train and predict the solutions. However, estimating such a loss function is not trivial.

Another limitation of ML models when predicting the solutions of optimisation problems is their limited capability for generalisation to other similar problems. For example, if a parameter is changed or a problem is applied to a different domain then one needs to train the model from scratch. However, some of the learning may be transferable to other problems. Transfer learning has been successfully applied in various RL and supervised ML problems \cite{brys2015policy}, but to date, there has not been any study in implementing transferring learning to optimisation problems. 

One way to improve the performance of ML in predicting the solutions of optimisation models is to use more rigorous feature engineering. In typical ML problems, the inputs have been the same as optimisation inputs, but this can be extended to include other features. For example, we can include features like descriptive statistics of data or variables related to the structure of the optimisation problems such as the number of constraints, and the number of violations of constraints to improve their predictability. \citet{Khalil_undated-qm} has summarised a set of variables that were used for ranking problems in a supervised ML model to choose the best branching variables. 

Last but not the least, the integration of predictive and prescriptive models requires software developers' attention. While there are many open-source software that can be used for either of these purposes, there is no software to predict the solutions of optimisation problems or to integrate prediction and optimisation.



\section{Conclusion}\label{sec:conclusion}

In this study, we explored the intersection of ML with prediction and optimisation. We discussed while prediction and optimisation have benefited from many theoretical advances and are largely applied for solving problems that are concerned with either of these methods, there is great power in integrating these two methods for solving real-world problems more efficiently. The rise of ML techniques and computational advances provides us with the best tools to integrate these methods and improve upon their performance.  We looked at two main paradigms on the intersection of ML, forecasting, and optimisation, namely \textit{predict and optimise}, and \textit{predicting the decisions of optimisation models}. We investigated several methodological advances and applications of each paradigm and discussed the current problems and future opportunities in research and practice. The research is still in its infancy in these paradigms, and there is a tremendous opportunity for researchers and practitioners. I hope this study can trigger ideas and future works in this domain.
\bibliography{biblio}

\end{document}